\documentclass[runningheads]{llncs}
\usepackage{enumitem}
 
\usepackage{eccv}

\usepackage{bigfoot}
\usepackage{eccvabbrv}

\usepackage{graphicx}
\usepackage{booktabs}
\usepackage{multirow}
\usepackage{xcolor}
\usepackage[accsupp]{axessibility}  

\usepackage[pagebackref,breaklinks,colorlinks,citecolor=eccvblue]{hyperref}

\usepackage{orcidlink}

\begin{document}

\title{EFF-Grasp: Energy-Field Flow Matching for Physics-Aware Dexterous Grasp Generation} 

\titlerunning{EFF-Grasp}

\author{Yukun Zhao\orcidlink{0009-0001-1431-9640} \and
Haoliang Sun\and
Zichen Zhong \and
Yongshun Gong \and
Yilong Yin}

\authorrunning{Y.~Zhao et al.}

\institute{Shandong University, Jinan, China\\
\email{haolsun@sdu.edu.cn}}

\maketitle

\begin{abstract}
Denoising generative models have recently become the dominant paradigm for dexterous grasp generation, owing to their ability to model complex grasp distributions from large-scale data. However, existing diffusion-based methods typically formulate generation as a stochastic differential equation (SDE), which often requires many sequential denoising steps and introduces trajectory instability that can lead to physically infeasible grasps. In this paper, we propose EFF-Grasp, a novel Flow-Matching-based framework for physics-aware dexterous grasp generation. Specifically, we reformulate grasp synthesis as a deterministic ordinary differential equation (ODE) process, which enables efficient and stable generation through smooth probability flows. To further enforce physical feasibility, we introduce a training-free physics-aware energy guidance strategy. Our method defines an energy-guided target distribution using adapted explicit physical energy functions that capture key grasp constraints, and estimates the corresponding guidance term via a local Monte Carlo approximation during inference. In this way, EFF-Grasp dynamically steers the generation trajectory toward physically feasible regions without requiring additional physics-based training or simulation feedback. Extensive experiments on five benchmark datasets show that EFF-Grasp achieves superior performance in grasp quality and physical feasibility, while requiring substantially fewer sampling steps than diffusion-based baselines.
\end{abstract}

\begin{figure}[h]
  \vspace{-12pt}
  \centering
  \includegraphics[width=0.7\linewidth]{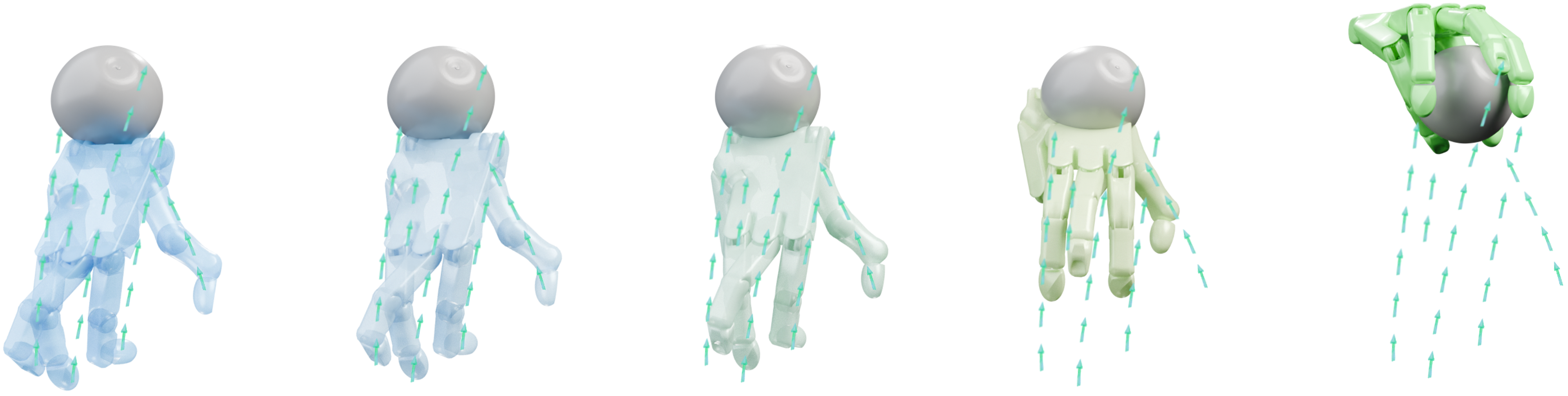}
  \caption{Visualization of the physics-aware generation process in EFF-Grasp.}
  \label{fig:intro}
  \vspace{-4pt}
\end{figure}

\section{Introduction}
\label{sec:intro}

Dexterous grasping, as a core capability in robotic manipulation, endows robots with the potential to interact with the complex physical world in a human-like manner. Unlike simple parallel-jaw grippers, multi-fingered hands possess high degrees of freedom (DoF), enabling versatile tasks but introducing the challenge of high-dimensional configuration spaces ($>20$ DoFs). Generating physically feasible grasps in such spaces efficiently remains an open problem.

To address this challenge, early analytical methods \cite{am1,am2,am3,am4,am5,am6} attempted to compute contact points through physical analysis but often suffered from the immense search space and complex optimization. More recently, data-driven methods have become the dominant paradigm. Among them, direct regression approaches \cite{ddg,DGTR} often produce averaged predictions and thus fail to capture the diversity of valid grasps. As a result, generative models, especially diffusion-based methods \cite{ddpm,karras2022elucidating,song2020sde,scenediffuser,lu2023ugg,weng2024dexdiffuser,zhang2024dexgrasp-diffusion,DGTR,GYS,zhong2025dexgrasp}, have gained increasing attention due to their strong ability to learn complex geometric patterns from large-scale data and synthesize diverse dexterous grasps. Despite their impressive performance, however, diffusion models inherently formulate generation as a SDE. This stochastic formulation introduces two important limitations. First, SDE-based sampling typically relies on many sequential denoising steps to progressively correct stochastic errors, which can limit sampling speed in practice \cite{song2023consistency}. Second, the injected noise at each sampling step causes trajectory jitter, making the generation process less stable and more likely to violate physical constraints \cite{albergo2025stochastic}. Therefore, the final grasp poses are more prone to penetration or non-contact.

In light of these limitations, we propose EFF-Grasp, a novel dexterous grasp generation framework that jointly improves sampling efficiency and physical feasibility. To address the inefficiency of SDE-based diffusion models, we reformulate dexterous grasp generation as a deterministic ODE process under the Flow Matching framework~\cite{lipman2022flow,liu2022flow,albergo2022building}. By learning a straight probability path between the noise and data distributions, our method eliminates per-step stochastic perturbations, enabling high-quality grasp generation with substantially fewer sampling steps, as illustrated in Fig. \ref{fig:intro}. To further ensure physical feasibility, we introduce a training-free physics-aware energy guidance strategy. Specifically, we define an energy-guided target distribution using adapted explicit physical energy functions tailored to dexterous grasp generation, which capture key grasp constraints, including external penetration, surface contact, and self-penetration. We then develop a local Monte Carlo approximation to estimate the corresponding guidance term during inference: at each generation step, the model predicts a candidate terminal grasp from the current state, samples local proposals around it, evaluates their physical energies, and uses these weighted samples to compute a physical energy field. This guidance is combined with the original Flow Matching velocity field, dynamically steering the trajectory toward physically feasible regions without requiring any additional physics-based training or simulation feedback. As a result, EFF-Grasp achieves efficient, robust, and physically grounded dexterous grasp generation within a unified framework.

In summary, the main contributions of this paper are as follows: 
\begin{itemize}[leftmargin=*,itemsep=0pt,topsep=2pt]
  \item[$\bullet$] We propose \textbf{EFF-Grasp}, a novel Flow-Matching-based framework for dexterous grasp generation, which formulates grasp synthesis as a deterministic ODE process and enables efficient and stable generation.
  
  \item[$\bullet$] We propose a \textbf{training-free physics-aware energy guidance} strategy for dexterous grasping. By incorporating adapted explicit physical energy fields (ERF, SPF, and SRF) with a local Monte Carlo approximation, our method enforces physical feasibility during inference without introducing additional training objectives or simulation-based optimization.
  
  \item[$\bullet$] We validate EFF-Grasp through extensive experiments on five benchmark datasets. EFF-Grasp achieves superior overall performance in grasp quality and physical realism, while significantly improving sampling efficiency compared with diffusion-based baselines.
\end{itemize}

\section{Method}
\label{sec:method}

\begin{figure}[t]
   \centering
  \includegraphics[width=\linewidth]{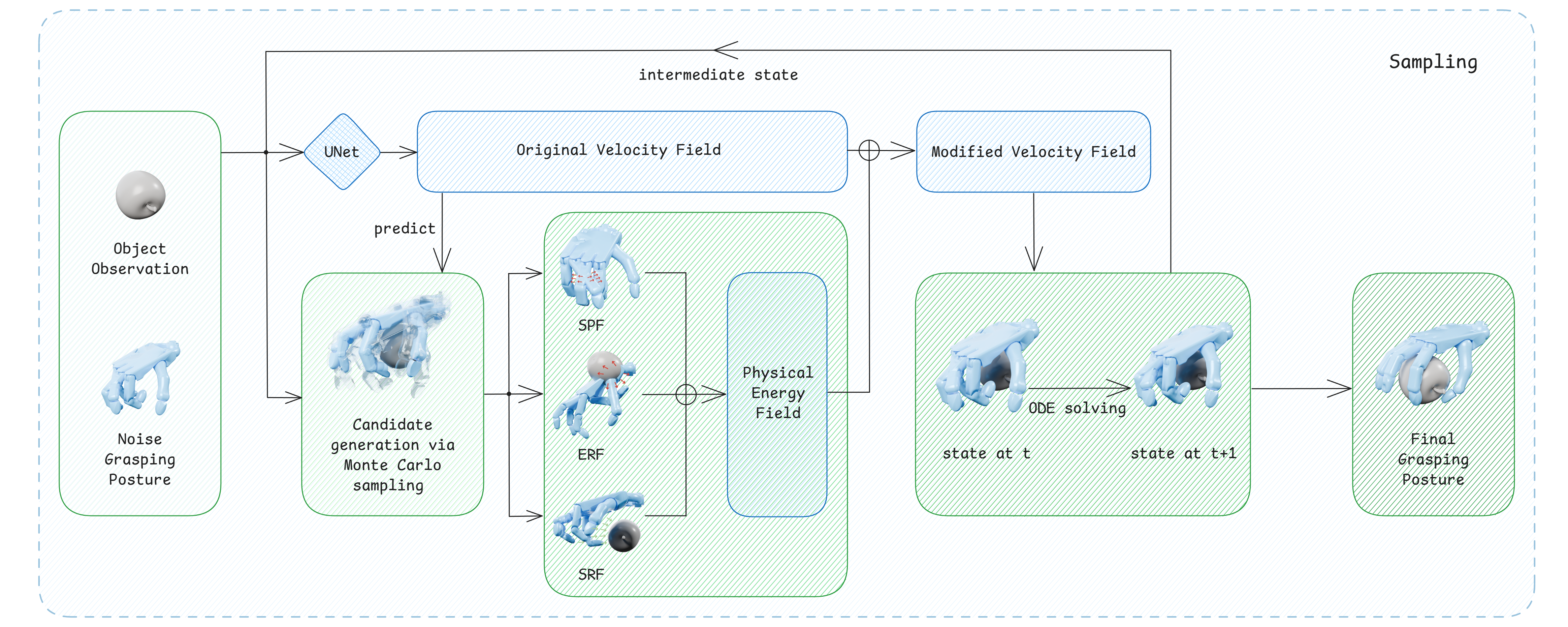}
  \caption{Overview of the EFF-Grasp sampling process. The method integrates Flow Matching with the training-free physics-aware guidance strategy. By evaluating candidate grasps using energy fields via local Monte Carlo sampling, we compute the physical energy field to guide the ODE trajectory towards physically feasible regions.}
  \label{fig:method}
\end{figure}

In this section, we introduce EFF-Grasp, a novel framework for dexterous grasp generation. As illustrated in Fig. \ref{fig:method}, we first formulate the dexterous grasping task and the Flow Matching training objective in Sec. \ref{sec:method:setup and train}. Subsequently, in Sec. \ref{sec:method:guidance}, we detail our Training-Free Physics-Aware Energy Guidance strategy, which integrates explicit physical energy fields to dynamically guide the generated trajectories towards physically feasible regions.

\subsection{Dexterous Grasp Generation via Flow Matching}
\label{sec:method:setup and train}

Our objective follows the standard setting of dexterous grasp generation: given an object, we seek to generate high-quality poses that can stably grasp it. Following the setting of DGA~\cite{zhong2025dexgrasp}, we define the dexterous hand state as $h = (\theta, R, t) \in \mathbb{R}^{D}$ (for ShadowHand, $D=33$). This state consists of three components: hand joint angles $\theta \in \mathbb{R}^{24}$, global rotation $R \in SO(3)$, and global translation vector $t \in \mathbb{R}^3$. Given a 3D object observation $O$ (typically represented as a point cloud), our goal is to sample a set of dexterous hand states $\mathcal{H} = \{h^{(i)}\}_{i=1}^N$ from the conditional distribution $p(h \mid O)$. 

Unlike prior methods that model the generation process as a SDE using Denoising Diffusion Probabilistic Models, we formulate the problem under the Flow Matching framework. Specifically, we aim to learn a time-dependent velocity field 
$v_t$, which defines the following ODE process:
\begin{equation}
  \frac{d h_t}{dt} = v_t(h_t, t, O), \quad h_0 \sim p_0 = \mathcal{N}(0, I), \quad h_1 \sim p_1(h \mid O),
  \label{eq:method:ode}
\end{equation}
where $t \in [0, 1]$ is the flow time variable. From this perspective, grasp pose generation is formulated as a continuous transport process that maps an initial standard Gaussian distribution $p_0$ to the target data distribution $p_1$ along the learned velocity field $v_t$.

To learn the velocity field $v_t$, we construct a probability path connecting the initial noise $p_0$ and the conditional data distribution $p_1(h\mid O)$. Unlike the complex stochastic paths in diffusion models, we employ a simple linear interpolation path. Specifically, for time step $t \in [0,1]$, the intermediate state $h_t$ is defined as:
\begin{equation}
  h_t = (1-(1-\sigma_{\min})t)\,h_0 + t\,h_1,
  \label{eq:method:ht}
\end{equation}
where $\sigma_{\min}$ is a minimal noise coefficient (e.g. $10^{-5}$) for numerical stability. Consequently, the corresponding target velocity field $u_t(h\mid h_1)$, which represents the time derivative of the path, is given by:
\begin{equation}
  u_t(h|h_1) = \frac{d h_t}{dt} = h_1 - (1-\sigma_{\min})h_0.
  \label{eq:method:ut}
\end{equation}
This linear path is not only computationally efficient but also yields straighter trajectories, which are easier for the model to fit.

To instantiate the model, we parameterize the velocity field $v_t$ with a neural network $v_{\theta}(h_t, t, O)$, typically implemented using a U-Net architecture. During training, the network is optimized to regress the target velocity field $u_t$ defined above. Concretely, the training objective is given by the mean squared error:
\begin{equation}
\mathcal{L}_{\mathrm{FM}} = \mathbb{E}_{t, h_0, h_1}\left[\left\lVert v_{\theta}(h_t, t, O) - u_t(h\mid h_1) \right\rVert_2^2\right]. 
\label{eq:method:fm-loss} 
\end{equation}

It is worth emphasizing that this training process does not incorporate any physical constraints; only a mean squared loss is required. In contrast to DGA~\cite{zhong2025dexgrasp}, which requires the introduction of complex physical gradients during training, our method relies on a \textbf{training-free sampling} strategy (see Sec. \ref{sec:method:guidance}). This design makes training both more stable and more efficient.

\subsection{Physical-Aware Sampling via Energy Field Flow}
\label{sec:method:guidance}

To ensure that the generated grasp poses not only conform to the data distribution $p_{1}$ but also strictly satisfy physical constraints (e.g., collision-free and tight contact), we model the generation target as an energy-guided distribution~\cite{levine2018reinforcement,janner2022planning}. Specifically, given a physical energy function $E(h, O)$, the ideal physically feasible distribution $p'(h|O)$ is defined in the Boltzmann form:
\begin{equation}
  p'(h\mid O) = \frac{1}{Z}\, p_1(h \mid O)\exp\!\left(-\frac{E(h, O)}{\tau}\right),
  \label{eq:method:boltzmann-target}
\end{equation}
where $p_1(h\mid O)$ is the primitive geometric distribution learned by the Flow Matching model, $\tau$ is a temperature parameter controlling the strength of physical constraints, and $Z=\int p_1(h)\exp(-E(h)/\tau)\,\text{d} h$ is the normalization constant.

To sample from this modified distribution $p'$, we leverage guidance techniques prevalent in diffusion models~\cite{dhariwal2021diffusion} and adapt the framework proposed by \cite{feng2025guidance} to our dexterous grasping task. Specifically, we augment the original velocity field $v_\theta$ with an additional guidance velocity field $g_t$. The resulting modified velocity governing the sampling trajectory is defined as:
\begin{equation}
  \hat{v}_t(h_t) = v_{\theta}(h_t) + s \cdot g_t(h_t),
  \label{eq:method:vhat-ht}
\end{equation}
where $s$ is the guidance scale. From a theoretical perspective, the optimal guidance term (energy field) is given by:
\begin{equation}
  g_t(h_t) = \int \left(\frac{e^{-E(h_0)/\tau}}{Z_t} - 1\right) u_{t \mid 0}(h_t \mid h_0)\, p(h_0 \mid h_t)\, \text{d} h_0,
  \label{eq:method:gt-theory}
\end{equation}
where $u_{t|0}$ is the conditional velocity field and $Z_t$ is a time-dependent normalization constant.

Although Eq. (\ref{eq:method:gt-theory}) provides an elegant theoretical form of the optimal guidance term, its practical application is hindered by two major challenges: i) {\textbf{how to define and compute an appropriate energy function for satisfying physical constraints}; ii) \textbf{how to evaluate the integral induced by this energy function, which is generally intractable for complex energy landscapes}. To address these challenges, we design an explicit physical guidance field and further introduce a local Monte Carlo approximation, making the guidance term practical.

\paragraph{\textbf{Explicit Physical Energy Fuctions.}}
To quantify the physical plausibility of a grasp pose $h$ given the object observation $O$, we construct a composite energy function $E(h, O)$, which consists of three components. Unlike previous works \cite{scenediffuser,xu2023unidexgrasp,zhong2025dexgrasp} that treat physical constraints as training losses, we utilize these potentials as energy guidance signals during the inference phase and \textbf{have made key improvements tailored for the energy-based sampling framework}.
\begin{enumerate}[label=\roman* ]
  \setlength{\itemsep}{2pt}
  \setlength{\topsep}{2pt}
  \setlength{\parskip}{0pt}
  \setlength{\parsep}{0pt}
\item \textit{External-Penetration Repulsion Field.}\cite{scenediffuser,zhong2025dexgrasp}
This potential aims to eliminate non-physical interpenetration between the hand and the object. Given the hand surface point set $P_{\mathrm{hand}}$ and the object point set $P_{\mathrm{obj}}$, we first compute the Signed Distance Function $\mathrm{SDF}(p_j, O)$ for each hand point $p_j \in P_{\mathrm{hand}}$ to the object surface (where negative values indicate penetration).

Existing methods typically use the average penetration depth as a penalty, which is not conducive to dilute severe local penetrations. To address this, we adopt a Max-Pooling strategy to heavily penalize the point with the most severe penetration:
\begin{equation}
  E_{\mathrm{ERF}}(h,O) = \max_{p_j \in P_{\mathrm{hand}}} \mathrm{ReLU}\!\left(-\mathrm{SDF}(p_j, O)\right).
  \label{eq:method:erf}
\end{equation}
This mechanism ensures that no part of the grasp configuration suffers from severe penetration, thus significantly enhancing physical feasibility.

  \item \textit{Surface Pulling Field.}\cite{xu2023unidexgrasp,zhong2025dexgrasp}
To generate grasps that fit tightly around the object, we need to encourage the fingers to approach the object surface. Existing contact constraints~\cite{xu2023unidexgrasp} often employ hard distance thresholds, applying attraction forces only to points within the threshold; this can easily lead to vanishing gradients when the hand is initialized far away. To solve this problem, we design a new two-level attraction potential containing global attraction and local refinement:
\begin{equation}
  E_{\mathrm{SPF}}(h,O) = \left\lVert c_{\mathrm{hand}} - c_{\mathrm{obj}} \right\rVert_2
  + \frac{1}{K} \sum_{p_k \in \mathrm{Top}\text{-}K} \left\lVert p_k - \mathrm{NN}(p_k, P_{\mathrm{obj}}) \right\rVert_2,
  \label{eq:method:spf}
\end{equation}
The first term calculates the Euclidean distance between the hand center $c_{\mathrm{hand}}$ and the object center $c_{\mathrm{obj}}$, providing long-range global attraction; the second term selects the $K$ keypoints closest to the object (Top-$K$) and calculates their average distance to the nearest neighbor points $\mathrm{NN}(p_k)$ on the object surface, serving for close-range contact refinement. This combined strategy ensures continuous and accurate guidance signals under both far and near initializations.

  \item \textit{Self-Penetration Repulsion Field.}\cite{scenediffuser,zhong2025dexgrasp}
To maintain the physical rationality of the hand structure, we need to prevent interpenetration between fingers~\cite{xu2023unidexgrasp}. We compute the pairwise Euclidean distances $d_{ij}$ between hand keypoints and penalize pairs with distances smaller than a safety threshold $\tau_{\mathrm{self}}$:
\begin{equation}
  E_{\mathrm{SRF}}(h,O) = \sum_{(i,j)\in \mathrm{Pairs}} \max\!\left(0, \tau_{\mathrm{self}} - d_{ij}\right).
  \label{eq:method:srf}
\end{equation}
\end{enumerate}

\noindent Combining the above three constraints, we define the total physical energy function as a weighted sum:
\begin{equation}
  E(h, O) = w_{\mathrm{ERF}} E_{\mathrm{ERF}}(h,O) + w_{\mathrm{SPF}} E_{\mathrm{SPF}}(h,O) + w_{\mathrm{SRF}} E_{\mathrm{SRF}}(h,O).
  \label{eq:method:energy-total}
\end{equation}
This energy function provides a quantitative measure of the physical infeasibility of a grasp pose $h$. A lower energy value indicates a higher likelihood that the grasp is physically feasible, thereby guiding the sampling process toward valid regions.

\paragraph{\textbf{Local Monte Carlo Estimation for $g_t$.}}
To approximate the theoretically optimal guidance derived in Eq. (\ref{eq:method:gt-theory}), we need to evaluate a high-dimensional integral involving the unknown posterior $p(h_0 \mid h_t)$. A direct approach is to approximate this integral via Monte Carlo sampling \cite{feng2025guidance}. Therefore, it is essential to design an efficient sampling strategy that can effectively leverage the physical energy field $E(h, O)$ defined above.

Targeting the dexterous grasping task, we exploit the linear trajectory property of the Flow Matching model and propose a tailored Monte Carlo estimation method. During inference, this method dynamically constructs a local proposal distribution and evaluates the resulting samples using analytic physical energy functions. The overall procedure consists of the following four steps:
\begin{enumerate}[label=\roman*]
  \setlength{\itemsep}{2pt}
  \setlength{\topsep}{2pt}
  \setlength{\parskip}{0pt}
  \setlength{\parsep}{0pt}
  
  \item \textit{Predict.} Utilizing the current model output $v_{\theta}(h_t)$, we estimate the potential final grasp pose by projecting along the tangent:
  \begin{equation}
    \hat{h}_{1|t} = h_t + (1-t) \cdot v_{\theta}(h_t).
    \label{eq:method:h1pred}
  \end{equation}

  \item \textit{Local Gaussian Exploration.} Given that $\hat{h}_{1|t}$ may not satisfy physical constraints, we construct a local Gaussian proposal distribution centered at $\hat{h}_{1|t}$ and sample $K$ candidate grasps:
  \begin{equation}
    h_1^{(k)} \sim \mathcal{N}\!\left(\hat{h}_{1|t}, \sigma_{\mathrm{local}}^2 I\right),\quad k=1,\dots,K,
    \label{eq:method:local-proposal}
  \end{equation}
  where $\sigma_{\mathrm{local}}$ controls the exploration range.

  \item \textit{Evaluate.} For each candidate solution, we compute the explicit physical energy $E(h_1^{(k)},O)$ and calculate the Boltzmann importance weight:
  \begin{equation}
    w_k = \exp\!\left(-\frac{E(h_1^{(k)},O)}{\tau}\right).
    \label{eq:method:weights}
  \end{equation}

  \item \textit{Guide.} Using Monte Carlo approximation, we convert the integral into a finite sample summation. Noting that the conditional velocity field pointing towards the target data is $u_{t|1}(h_t \mid  h_1^{(k)})=(h_1^{(k)} - h_t)/(1-t)$, the final guidance velocity field $g_t$ is calculated as:
  \begin{equation}
    g_t(h_t) \approx \frac{1}{K} \sum_{k=1}^{K} \left( \frac{w_k}{\bar{w}} - 1 \right) \frac{h_1^{(k)} - h_t}{1-t},
    \label{eq:method:gt-mc}
  \end{equation}
  where $\bar{w}=\frac{1}{K}\sum_{k=1}^{K} w_k$ is the mean weight. Intuitively, this term pulls the generation trajectory towards the direction of physically feasible grasp candidates.
\end{enumerate}

\noindent Through the proposed strategy, physical deviations are dynamically detected and corrected at each generation step, thereby enabling training-free satisfaction of physical constraints.
\section{Experiments}
\label{sec:experiments}

In this section, we conduct extensive experiments to validate the performance of EFF-Grasp. We aim to answer the following three core questions:

\begin{enumerate} [label=\roman*]
\item \textbf{Generation Quality and Physical Feasibility.} Compared to existing diffusion-based methods, can our method significantly reduce physical penetration and improve grasp success rates without requiring complex fine-tuning?
\item \textbf{Inference Efficiency.} Can EFF-Grasp achieve a superior trade-off between quality and efficiency among training-free methods? Specifically, how does the linear trajectory of Flow Matching contribute to efficient sampling compared to stochastic diffusion baselines?
\item \textbf{Effectiveness of Guidance Mechanism.} Does the improved physical energy guidance (ERF, SPF, SRF) outperform the original version? What are the specific contributions of each component to the final performance? 
\end{enumerate}

\begin{table*}[t]
  \centering
  \caption{Quantitative comparison on three large-scale datasets. Best results are in \textbf{bold}, and second-best are \underline{underlined}.}
  \label{tab:large-scale-results}
  \setlength{\tabcolsep}{3.5pt}
  \scriptsize
  
  \resizebox{\textwidth}{!}{%
    \begin{tabular}{l cccc cccc cccc}
      \toprule
      \multirow{2}{*}{\textbf{Method}} & \multicolumn{4}{c }{\textbf{DexGraspNet}} & \multicolumn{4}{c }{\textbf{UniDexGrasp}} & \multicolumn{4}{c}{\textbf{DexGRAB}} \\
      \cmidrule(lr){2-5}\cmidrule(lr){6-9}\cmidrule(lr){10-13}
        & Suc.6$\uparrow$ & Suc.1$\uparrow$ & Pen.$\downarrow$ & Div$\uparrow$
        & Suc.6$\uparrow$ & Suc.1$\uparrow$ & Pen.$\downarrow$ & Div$\uparrow$
        & Suc.6$\uparrow$ & Suc.1$\uparrow$ & Pen.$\downarrow$ & Div$\uparrow$ \\
      \midrule
      UniDexGrasp~\cite{xu2023unidexgrasp} & 33.9 & 70.1 & 31.9 & 0.14 & 23.7 & 65.5 & 24.5 & 0.14 & 20.8 & 55.8 & 37.4 & 0.08 \\
      GraspTTA~\cite{grasptta}    & 18.6 & 67.8 & 24.5 & 0.13 & 21.0 & 65.3 & 21.2 & 0.10 & 14.4 & 51.0 & 51.4 & 0.10 \\
      SceneDiffuser~\cite{scenediffuser}& 26.6 & 66.9 & 31.0 & 0.15 & 28.3 & 74.8 & 25.1 & \underline{0.15} & 39.1 & 85.0 & 41.1 & \textbf{0.12} \\
      UGG~\cite{lu2023ugg}          & 46.9 & 79.0 & 25.2 & 0.14 & 46.0 & 83.2 & 24.5 & 0.14 & 42.7 & 90.6 & 33.2 & \textbf{0.12} \\
      \midrule
      DGA~\cite{zhong2025dexgrasp}          & \underline{53.6} & \underline{90.4} & \underline{21.5} & \textbf{0.22} & \underline{54.8} & \underline{90.8} & \textbf{18.9} & \textbf{0.25} & \underline{56.5} & \underline{91.8} & \underline{28.6} & \textbf{0.12} \\
      \textbf{EFF-Grasp (Ours)}             & \textbf{67.2} & \textbf{91.2} & \textbf{20.7} & \underline{0.16} & \textbf{68.0} & \textbf{92.6} & \underline{19.3} & \underline{0.15} & \textbf{61.3} & \textbf{91.9} & \textbf{23.6} & \underline{0.11} \\      
      \bottomrule
    \end{tabular}%
  }
\end{table*}

\begin{table*}[t]
  \centering
  \caption{Quantitative comparison on small-scale and real-world datasets. Best results are in \textbf{bold}, and second-best are \underline{underlined}.}
  \label{tab:small-scale-results}
  \setlength{\tabcolsep}{3.5pt}
  \scriptsize
  
  \resizebox{0.7\textwidth}{!}{%
    \begin{tabular}{l cccc cccc}
      \toprule
      \multirow{2}{*}{\textbf{Method}} & \multicolumn{4}{c}{\textbf{MultiDex}} & \multicolumn{4}{c}{\textbf{RealDex}} \\
      \cmidrule(lr){2-5}\cmidrule(lr){6-9}
        & Suc.6$\uparrow$ & Suc.1$\uparrow$ & Pen.$\downarrow$ & Div$\uparrow$
        & Suc.6$\uparrow$ & Suc.1$\uparrow$ & Pen.$\downarrow$ & Div$\uparrow$ \\
      \midrule
      UniDexGrasp~\cite{xu2023unidexgrasp} & 21.6 & 47.5 & 13.5 & 0.08 & 27.1 & 59.4 & 39.0 & \underline{0.11} \\
      GraspTTA~\cite{grasptta}    & 30.3 & 62.8 & 19.0 & 0.11 & 13.3 & 46.4 & 40.1 & 0.09 \\
      SceneDiffuser~\cite{scenediffuser}& 69.8 & 85.6 & 14.6 & \textbf{0.27} & 21.7 & 56.1 & 42.0 & 0.09 \\
      UGG~\cite{lu2023ugg}          & 55.3 & 93.4 & \underline{10.3} & 0.12 & 32.7 & 63.4 & 34.4 & 0.10 \\
      \midrule
      DGA~\cite{zhong2025dexgrasp}          & \underline{72.2} & \underline{96.3} & \textbf{9.6} & \underline{0.23} & \underline{34.6} & \underline{71.2} & \underline{23.1} & \textbf{0.14} \\
      \textbf{EFF-Grasp (Ours)}             & \textbf{77.5} & \textbf{97.5} & 10.5 & 0.13 & \textbf{41.0} & \textbf{74.2} & \textbf{21.9} & 0.10 \\      
      \bottomrule
    \end{tabular}%
  }
\end{table*}

\subsection{Experimental Setup}
\label{sec:exp:setup}

\subsubsection{Datasets.}
To comprehensively evaluate the generalization ability of our model, we conduct experiments on five challenging dexterous grasping datasets: DexGraspNet\cite{wang2023dexgraspnet}, UniDexGrasp\cite{xu2023unidexgrasp}, DexGRAB\cite{taheri2020grab}, RealDex\cite{liu2024realdex}, and MultiDex\cite{gendexgrasp-multidex}. The first three datasets are large-scale in size and scope, e.g., DexGraspNet . We use the official train/test splits for all experiments.

\subsubsection{Evaluation Metrics.}
Following the standard evaluation protocols of SceneDiffuser\cite{scenediffuser} and DGA\cite{zhong2025dexgrasp}, we conduct physical simulation tests in the Isaac Gym\cite{makoviychuk2021isaac} simulator and use the following metrics to measure generation quality: 
\begin{itemize}
    \item \textit{Success Rate} (Suc.$\uparrow$): The core metric for evaluating grasp stability. A grasp is considered successful if the object displacement is less than 2\,cm after applying external forces. We report Suc.1 (stable under at least one random external force) and Suc.6 (stable under six orthogonal external forces).
    \item \textit{Maximum Penetration Depth} (Pen.$\downarrow$): Measures the physical realism of the generated results. It calculates the maximum penetration distance (mm) between the hand and object meshes; lower is better.
    \item \textit{Diversity} (Div.$\uparrow$): Measures the coverage of the generated distribution. It calculates the average standard deviation of successful grasp poses in the joint space; higher indicates wider coverage.
\end{itemize}

\subsubsection{Baselines.}
To comprehensively evaluate performance, we compare EFF-Grasp with a range of advanced methods, including GraspTTA \cite{grasptta}, and generative approaches such as UniDexGrasp \cite{xu2023unidexgrasp}, SceneDiffuser \cite{scenediffuser}, UGG \cite{lu2023ugg}, and the recent SOTA diffusion model DexGrasp Anything \cite{zhong2025dexgrasp}. Unlike these methods which often rely on complex training strategies, our EFF-Grasp achieves superior performance in a completely training-free guidance strategy.

\subsubsection*{Implementation Details.}
Similar to \cite{scenediffuser,zhong2025dexgrasp}, we adapt an identical U-Net \cite{ronneberger2015u} tailored for the Flow Matching architecture as our backbone network. We use the Adam optimizer\cite{kingma2014adam} with a learning rate of $1e-4$ and train on a Linux server equipped with 4 NVIDIA GeForce RTX™ 3090 GPUs until convergence. Regarding sampling settings, baseline methods (e.g., DGA) use their default 100 steps; our EFF-Grasp also uses 100 steps in the main experiments, while we vary the sampling steps in the efficiency experiments to demonstrate its advantages. For the proposed physics-aware energy guidance, we empirically determined the optimal hyperparameters: the weights for the three energy fields are all set to $0.4$, the guidance scale $s$ is set to $30$, and the temperature $\tau$ is $0.05$.

\begin{figure}[t]
   \centering
  \includegraphics[width=\linewidth]{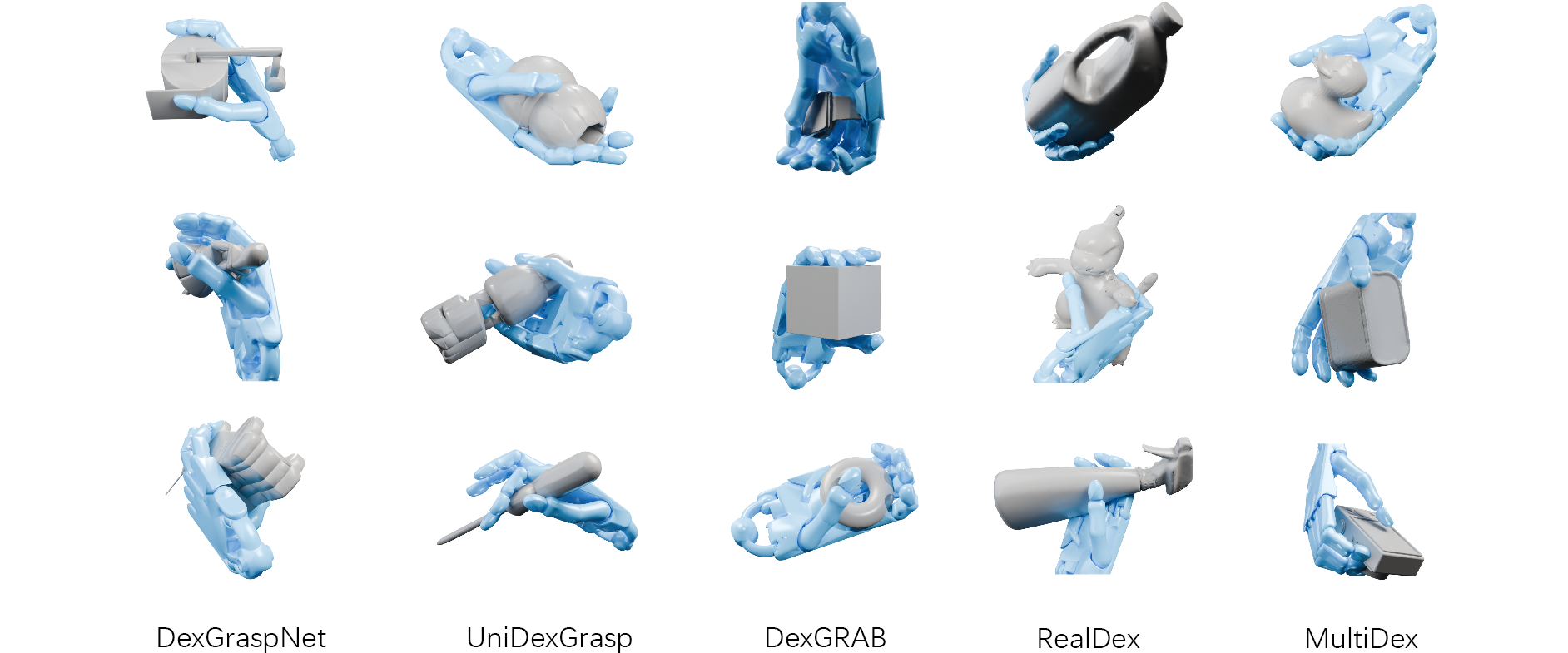}
  \caption{Visualization of generation results produced by our method across all datasets.}
  \label{fig:result}
\end{figure}

\subsection{Comparative Analysis of Grasp Generation Performance} 
\label{sec:exp:comparative-analysis}

\subsubsection{Overall Performance and Physical Realism.}
\label{sec:exp:overall-performance}

\begin{figure}[t]
  \centering
  \includegraphics[width=\linewidth]{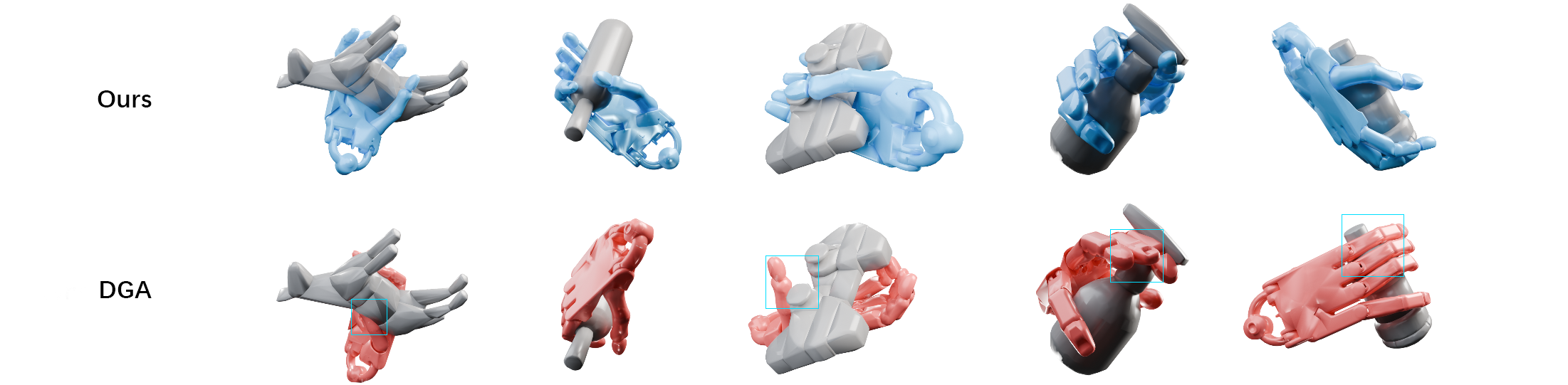}
  \caption{Compared to the SOTA method DGA, EFF-Grasp generates more stable and physically feasible grasps (see the boxed failure cases of DGA).}
  \label{fig:qualitative-4-2}
  \vspace{-8pt}
\end{figure}

To answer the first question, we report the quantitative comparison between EFF-Grasp and all baselines in Tab. \ref{tab:large-scale-results} \& \ref{tab:small-scale-results}. The results show that our method outperforms current state-of-the-art approaches in both generation quality and physical realism.

Compared with earlier generative models, including GraspTTA, UniDexGrasp, and SceneDiffuser, EFF-Grasp shows clear advantages. Even against the strongest baseline, DGA, our method improves Suc.6 on the challenging DexGraspNet dataset from 53.6\% to 67.2\%, a gain of +13.6\%. On real-world datasets such as RealDex and DexGRAB, Suc.6 further improves by +6.5\% and +4.8\%, respectively. These results indicate that our method better captures the grasp distribution over complex geometric surfaces.

In terms of physical feasibility, our method achieves lower Pen. on most datasets. In particular, on DexGRAB, the penetration depth is reduced from 28.6mm for DGA to 23.6mm. This confirms that our physical energy guidance, especially ERF and SPF, effectively steers generation away from physically infeasible regions and produces cleaner, more physically plausible grasps. Fig. \ref{fig:result} presents the visualized grasping results of this superior grasp.

\subsubsection{Stability vs. Diversity Analysis.} 
\label{sec:exp:tradeoff}

Although EFF-Grasp achieves the best overall performance in success rate and physical metrics, it is slightly lower than DGA on the Div. metric in Tab. \ref{tab:large-scale-results} \& \ref{tab:small-scale-results}. We argue that this does not indicate a limitation of the model, but instead reflects an inherent \textbf{stability-diversity balance} arising from different generation paradigms, namely ODE versus SDE.

Specifically, DGA is based on the DDPM, whose SDE-based sampling process injects random noise at every denoising step. While this stochasticity promotes broader exploration and thus higher diversity, it also increases the chance of deviating from the physical manifold, resulting in floating or penetrating grasps (see Fig. \ref{fig:qualitative-4-2}). By contrast, EFF-Grasp adopts Flow Matching, whose ODE-based sampling yields smoother and more deterministic trajectories once the initial noise is fixed. This makes the model more likely to converge to stable and physically feasible grasp modes, rather than pursuing marginal solutions for additional diversity. In dexterous grasping, where success rate and physical realism are of primary importance, this trade-off is both reasonable and beneficial.

\subsection{Efficiency and Quality Analysis}
\label{sec:exp:efficiency}

\begin{figure}[t]
   \centering
  \includegraphics[width=\linewidth]{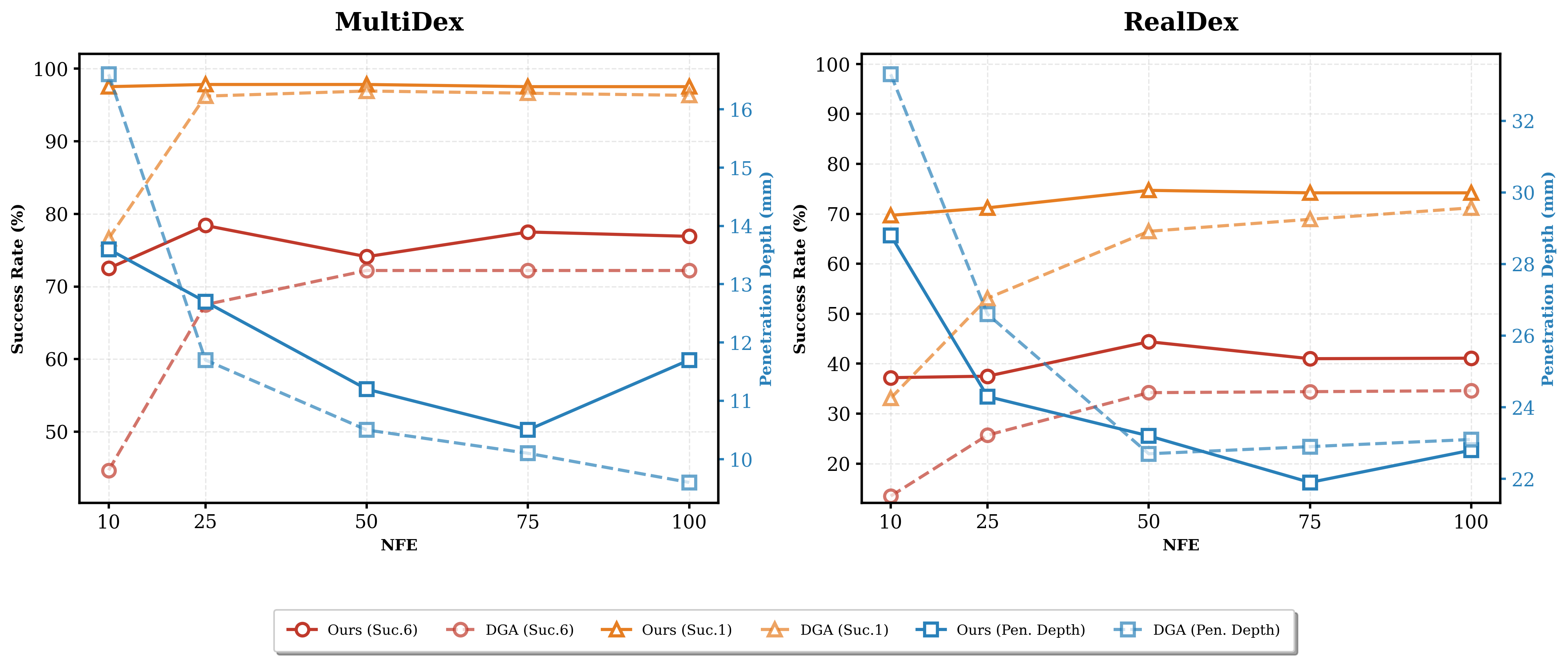}
  \caption{Comparison of Success Rate and Penetration Depth versus NFE on MultiDex and RealDex datasets. The left y-axis represents the Success Rate (\%), and the right y-axis indicates the Penetration Depth (mm). Solid lines denote our method, while dashed lines represent the DGA baseline. Our method consistently outperforms DGA, achieving significantly higher success rates and lower penetration depths, particularly demonstrating faster convergence at lower NFEs.}
  \label{fig:nfe-efficiency}
  \vspace{-8pt}
\end{figure}

To answer the second question, we study the effect of different sampling steps (NFE) on generation quality. Diffusion-based methods typically require many sampling steps to converge, which limits their efficiency in real-time planning.

We evaluate the performance of both algorithms under NFE settings ranging from 10 to 100. As shown in Fig. \ref{fig:nfe-efficiency}, EFF-Grasp is highly efficient due to the smooth ODE trajectory of Flow Matching. Unlike diffusion-based approaches, even at very low NFE (e.g., 10), the model still maintains a usable grasp structure without drastic degradation. This enables EFF-Grasp to flexibly adjust sampling cost under different computational budgets while preserving generation quality, demonstrating the efficiency advantage of ODE-based Flow Matching over SDE-based diffusion models.

\subsection{Ablation Study}
\label{sec:exp:ablation}

To answer the third question, we perform a comprehensive ablation study on the DexGraspNet dataset to evaluate the individual contributions of our physical guidance modules and the effectiveness of the adapted energy functions.

\textbf{Setup.}
We adopt a controlled ablation setup, starting from a baseline model with vanilla Flow Matching generation. We then progressively incorporate the three redesigned physical energy functions: SRF, ERF, and SPF. In addition, to verify the advantage of our explicit energy function design, we implement a variant that applies the original forms of SRF, ERF, and SPF as guidance within the same Flow Matching framework. All variants are evaluated using the same metrics as in the main experiments.

\begin{table*}[t]
  \centering
  \caption{Ablation study of physical energy functions on DexGraspNet.}
  \label{tab:ablation-physics-guidance}
  \setlength{\tabcolsep}{4pt}
  \scriptsize
  \begin{tabular}{c l c c c c c c c}
    \toprule
    ID & Model Variant & SRF & ERF & SPF & Suc.6$\uparrow$ & Suc.1$\uparrow$ & Pen.$\downarrow$ & Div.$\uparrow$ \\
    \midrule
    a & Baseline (Vanilla FM) &  &  &  & 47.0 & 90.7 & 17.1 & 0.16 \\
    b & + SRF & $\checkmark$ &  &  & 48.0 & 91.2 & 16.0 & 0.16 \\
    c & + ERF &  & $\checkmark$ &  & 45.5 & 90.4 & \textbf{14.7} & 0.15 \\
    d & + SRF \& ERF & $\checkmark$ & $\checkmark$ &  & 47.1 & 90.8 & 15.4 & 0.15 \\
    e & + SRF \& ERF \& SPF & $\checkmark$ & $\checkmark$ & $\checkmark$ & \textbf{67.2} & \textbf{91.2} & 20.7 & 0.16 \\
    f & + SRF \& ERF \& SPF (Original forms) & $\checkmark$ & $\checkmark$ & $\checkmark$ & 66.7 & 91.1 & 20.7 & 0.16 \\
    \bottomrule
  \end{tabular}
\end{table*}

\begin{figure}[t]
  \centering
  \includegraphics[width=\linewidth]{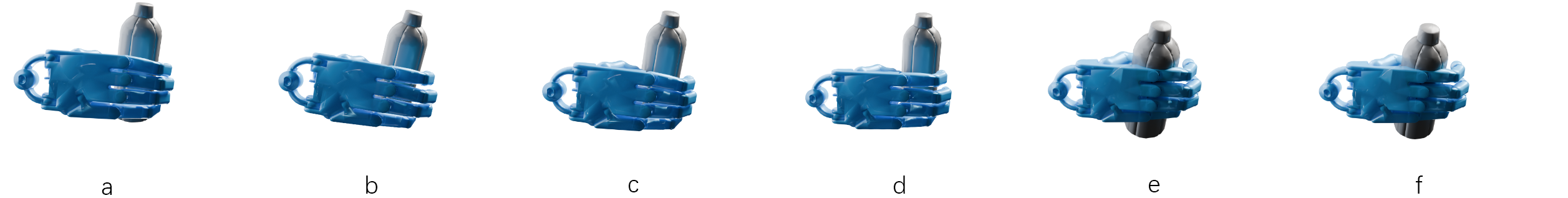}
  \caption{Qualitative visualization of the ablation study. Variant e yields the optimal results (bolded in Table \ref{tab:ablation-physics-guidance}), validating the effectiveness of our proposed energy terms over the baseline (a) and original forms (f).}
  \label{fig:qualitative-4-4}
  \vspace{-8pt}
\end{figure}

\textbf{Analysis.}
The quantitative results reported in Table \ref{tab:ablation-physics-guidance} and Fig. \ref{fig:qualitative-4-4} highlight the distinct roles of each module: SRF and ERF mainly regularize hand poses and reduce interpenetration, whereas SPF is the key to grasp stability. Specifically, adding SRF (Line b) encourages more natural hand poses, resulting in slight improvements in both success rate and penetration depth. Incorporating ERF alone (Line c) further reduces the penetration depth to the lowest level (14.7mm), but at the cost of a small drop in Suc.6 (47.0\% $\rightarrow$ 45.5\%). This behavior is consistent with physical intuition: ERF repels fingers from the object to avoid collisions, but such a conservative strategy can also produce looser grasps that are less robust to external forces. In contrast, introducing SPF (Line e) leads to a substantial performance gain, with Suc.6 increasing sharply from 47.1\% to 67.2\%. The contact term in SPF encourages the fingers to form tight and valid contact with the object surface, thereby significantly improving friction and grasp stability. Although this comes with a moderate increase in penetration, the trade-off is reasonable and necessary for achieving high grasp success.

\textbf{Effectiveness of Redesigned Energy Functions.}
Compared to Line f (original design), Line e (our improved functions) achieves consistently higher Suc. rates, demonstrating that the redesigned energy functions more effectively guide the Flow Matching sampler toward higher-quality grasps.

\subsection{Cross-Dataset Generalization}
\label{sec:exp:cross-dataset}

To assess the generalization capability of EFF-Grasp, we perform cross-dataset validation experiments. Specifically, the model is trained on dataset A and directly evaluated on an unseen dataset B without any fine-tuning. The results are reported in Tab.~\ref{tab:cross-dataset}. EFF-Grasp demonstrates consistently strong cross-dataset generalization across most dataset pairs. These results demonstrate the high efficiency of the proposed sampling method.

\begin{table*}[t]
  \centering
  \caption{EFF-Grasp achieves superior cross-dataset generalization performance.}
  \label{tab:cross-dataset}
  \resizebox{\linewidth}{!}{
  \begin{tabular}{ccc|ccc|ccc|ccc}
    \toprule
    \makebox[1.5cm][c]{Suc.6$\uparrow$} & \makebox[1.5cm][c]{Suc.1$\uparrow$} & \makebox[1.5cm][c]{Pen.$\downarrow$} & \makebox[1.5cm][c]{Suc.6$\uparrow$} & \makebox[1.5cm][c]{Suc.1$\uparrow$} & \makebox[1.5cm][c]{Pen.$\downarrow$} & \makebox[1.5cm][c]{Suc.6$\uparrow$} & \makebox[1.5cm][c]{Suc.1$\uparrow$} & \makebox[1.5cm][c]{Pen.$\downarrow$} & \makebox[1.5cm][c]{Suc.6$\uparrow$} & \makebox[1.5cm][c]{Suc.1$\uparrow$} & \makebox[1.5cm][c]{Pen.$\downarrow$} \\
    \midrule
    \multicolumn{3}{c|}{\footnotesize \textbf{DexGraspNet $\to$ UniDexGrasp}} & \multicolumn{3}{c|}{\footnotesize \textbf{DexGraspNet $\to$ DexGRAB}} & \multicolumn{3}{c|}{\footnotesize \textbf{DexGraspNet $\to$ RealDex}} & \multicolumn{3}{c}{\footnotesize \textbf{DexGraspNet $\to$ MultiDex}} \\
    70.0& 93.5 & 19.8 & 60.8 & 91.0 & 23.8 & 41.0 & 80.1 & 18.6 & 61.0 & 93.4 & 8.0 \\
    \midrule
    \multicolumn{3}{c|}{\footnotesize \textbf{UniDexGrasp $\to$ DexGraspNet}} & \multicolumn{3}{c|}{\footnotesize \textbf{UniDexGrasp $\to$ DexGRAB}} & \multicolumn{3}{c|}{\footnotesize \textbf{UniDexGrasp $\to$ RealDex}} & \multicolumn{3}{c}{\footnotesize \textbf{UniDexGrasp $\to$ MultiDex}} \\
    67.2 & 91.5 & 21.6 & 60.6 & 91.7 & 24.1 & 42.8 & 78.8 & 20.2 & 59.4 & 94.1 & 9.5 \\
    \midrule
    \multicolumn{3}{c|}{\footnotesize \textbf{DexGRAB $\to$ RealDex}} & \multicolumn{3}{c|}{\footnotesize \textbf{DexGRAB $\to$ MultiDex}} & \multicolumn{3}{c|}{\footnotesize \textbf{RealDex $\to$ DexGRAB}} & \multicolumn{3}{c}{\footnotesize \textbf{RealDex $\to$ MultiDex}} \\
    29.0 & 58.5 & 26.1 & 55.0 & 94.7 & 9.6 & 61.0 & 91.3 & 23.5 & 51.9 & 90.0 & 8.3 \\
    \bottomrule
  \end{tabular}
  }\vspace{-8pt}
\end{table*}
\section{Related Work}
\label{sec:related}

\subsubsection{Dexterous Grasp Generation.}
This task aims to synthesize stable and physically feasible grasp poses for multi-fingered hands. Research in this area has evolved from analytical methods to data-driven approaches. Early analytical methods~\cite{am1,am2,am3,am4,am5,am6} relied on physical models and force-closure analysis, but struggled with high-dimensional search spaces and unknown objects, as their computational complexity grows exponentially with the problem size.

With the rise of deep learning, data-driven methods gradually became the dominant paradigm. Early regression-based methods~\cite{ddg,DGTR} typically produced a single averaged prediction, and therefore failed to capture the inherent diversity of valid grasps. Subsequent generative methods~\cite{bousmalis2018using,grasptta,xu2023unidexgrasp} began to model grasp distributions explicitly, but their generation quality remained limited. More recently, diffusion-based methods~\cite{scenediffuser,lu2023ugg,zhong2025dexgrasp} have achieved state-of-the-art performance thanks to their strong distribution modeling ability. However, these methods are usually based on stochastic differential equations (SDEs), requiring many sampling steps during denoising and thus suffering from low inference efficiency. Moreover, the stochasticity of SDE-based sampling can introduce trajectory jitter, which increases the risk of physical interpenetration. Although EvolvingGrasp~\cite{zhu2025evolvinggrasp} improves efficiency through Consistency Models~\cite{kim2023consistency,lu2024simplifying}, it still depends on an additional distillation process during training.

Recently, Flow Matching has also been explored in robotic manipulation. For example, DemoGrasp~\cite{yuan2025demograsp} learns universal grasping actions from demonstrations. However, these methods mainly focus on learning policy or action distributions from data, without explicitly modeling physical constraints or supporting training-free physical adaptation at inference time. In contrast, our EFF-Grasp adopts a deterministic Flow Matching framework, which \textbf{alleviates the stochasticity and efficiency limitations of SDE-based methods, while further ensuring physical feasibility through explicit energy guidance.}

\subsubsection{Physics-Aware Grasp Synthesis.}
Ensuring physical feasibility remains a central challenge in dexterous grasp generation. Traditional analytical methods attempt to optimize contact energy directly, but they often suffer from the curse of dimensionality. More recent learning-based methods seek to incorporate physical constraints into generative models. For example, DGA~\cite{zhong2025dexgrasp} introduces physical losses during training and further applies physical gradients for guidance at inference time. However, under its SDE-based generation paradigm, the injected random noise can interfere with gradient guidance, leading to trajectory oscillation. Another line of work relies on preference alignment or reinforcement learning. For instance, EvolvingGrasp~\cite{zhu2025evolvinggrasp} directly fine-tunes the model using simulation feedback and achieves strong performance. Nevertheless, this generation-simulation-finetuning loop is computationally expensive and heavily dependent on the evaluation simulation environment. In contrast, EFF-Grasp adopts a novel training-free physics-aware energy guidance strategy. By directly leveraging explicit physical energy fields to guide the deterministic Flow Matching trajectory, our method \textbf{avoids complex training pipelines while enabling efficient and flexible generation of high-quality, physically feasible grasps}.

\subsubsection{Flow Matching and Trajectory Guidance.}
Flow Matching \cite{lipman2022flow,liu2022flow,albergo2022building} has emerged as a powerful generative framework that learns vector fields along predefined conditional probability paths. Compared with diffusion models based on SDEs and normalizing-flow-based grasp synthesis methods \cite{feng2024ffhflow}, it defines a deterministic continuous transformation from noise to data by solving an ODE, offering advantages in both sampling efficiency and generation quality.

Guidance is equally important for conditional generation. In diffusion models, common approaches such as classifier guidance~\cite{dhariwal2021diffusion} and reconstruction-based guidance like DPS~\cite{chung2022diffusion} typically depend on pretrained classifiers or expensive gradient computations. More recently, flow guidance theory~\cite{feng2025guidance} has provided a general energy-based framework. However, extending these ideas to dexterous grasping is challenging, since existing work mainly targets image generation or low-dimensional geometry and does not model complex physical interactions such as contact and interpenetration. In this work, we extend these theoretical foundations to dexterous grasp generation and, together with our redesigned explicit physical energy fields (ERF, SPF, and SRF), propose a physics-aware energy guidance strategy that enables efficient, training-free, and physically constrained generation within the Flow Matching framework.}
 
\section{Conclusion}
\label{sec:conclusion}

We presented EFF-Grasp, a simple yet effective physics-aware generative framework built on energy-guided Flow Matching. Through a training-free guidance mechanism, it enables universal and high-quality dexterous grasp generation with low sampling cost. By leveraging the straight-line trajectory property of Flow Matching, our method substantially reduces the number of sampling steps, while the proposed physical energy guidance further improves grasp quality and physical feasibility. We believe that the efficiency and practicality of EFF-Grasp make it a promising solution for deploying dexterous grasping in real-world industrial scenarios.


%
%

\bibliographystyle{splncs04}
\bibliography{main}

\newpage
\newpage\appendix

\section{Derivation of Final Pose Prediction in Eq. (\ref{eq:method:h1pred})}
\label{sec:appendix:derivation}

\paragraph{Motivation.}
In our proposed energy guidance framework, we require evaluating the physical feasibility of the generated grasps using the energy function $E(h)$. However, during the intermediate steps of the Flow Matching process ($t < 1$), the state $h_t$ represents a linear interpolation between Gaussian noise and the grasp pose. Applying physical constraints directly on this noisy intermediate state is physically meaningless, as the geometric structures are not yet formed. Therefore, we cannot directly compute energy on $h_t$. Instead, we must estimate the potential final clean grasp pose $\hat{h}_{1|t}$ implied by the current trajectory, and evaluate the energy on this predicted pose. This necessitates a formula to recover $\hat{h}_{1|t}$ from the current state $h_t$ and velocity $v_{\theta}(h_t)$.

\paragraph{Derivation.}
In this section, we provide the detailed derivation for this final grasp pose estimation formula (Eq. \ref{eq:method:h1pred}) used in our strategy.

Recall that in the Flow Matching framework Sec. \ref{sec:method:setup and train}, we define the probability path from the source distribution (noise) to the target distribution (data) using a linear interpolation. Specifically, the intermediate state $h_t$ at time $t$ is defined as:
\begin{equation}
  h_t = \alpha_t h_0 + \beta_t h_1,
  \label{eq:appendix:ht}
\end{equation}
where $h_0 \sim \mathcal{N}(0, I)$ is the initial noise, $h_1$ is the target grasp pose, and the coefficients are given by $\alpha_t = 1 - (1-\sigma_{\min})t$ and $\beta_t = t$. Here, $\sigma_{\min}$ is a small constant for numerical stability.

The corresponding target velocity field $u_t$, which the neural network $v_{\theta}(h_t, t)$ aims to learn, is the time derivative of the path:
\begin{equation}
  u_t = \frac{d h_t}{dt} = \dot{\alpha}_t h_0 + \dot{\beta}_t h_1,
  \label{eq:appendix:ut}
\end{equation}
where $\dot{\alpha}_t = -(1-\sigma_{\min})$ and $\dot{\beta}_t = 1$.

During inference, we aim to estimate the unknown target pose $h_1$ given the current state $h_t$ and the predicted velocity $v_{\theta}(h_t) \approx u_t$. We can view Eqs. \eqref{eq:appendix:ht} and \eqref{eq:appendix:ut} as a linear system with unknowns $h_0$ and $h_1$:
\begin{equation}
  \begin{cases}
    h_t = \alpha_t h_0 + \beta_t h_1 \\
    v_{\theta}(h_t) \approx \dot{\alpha}_t h_0 + \dot{\beta}_t h_1
  \end{cases}
\end{equation}

To solve for $h_1$, we eliminate $h_0$. Multiplying the first equation by $\dot{\alpha}_t$ and the second by $\alpha_t$, we get:
\begin{equation}
  \begin{cases}
    \dot{\alpha}_t h_t = \dot{\alpha}_t \alpha_t h_0 + \dot{\alpha}_t \beta_t h_1 \\
    \alpha_t v_{\theta}(h_t) \approx \alpha_t \dot{\alpha}_t h_0 + \alpha_t \dot{\beta}_t h_1
  \end{cases}
\end{equation}

Subtracting the first equation from the second eliminates $h_0$:
\begin{equation}
  \alpha_t v_{\theta}(h_t) - \dot{\alpha}_t h_t \approx (\alpha_t \dot{\beta}_t - \dot{\alpha}_t \beta_t) h_1.
\end{equation}

Thus, the estimated final pose $\hat{h}_{1|t}$ is given by:
\begin{equation}
  \hat{h}_{1|t} = \frac{\alpha_t v_{\theta}(h_t) - \dot{\alpha}_t h_t}{\alpha_t \dot{\beta}_t - \dot{\alpha}_t \beta_t}.
\end{equation}

Substituting the specific values for our linear path:
\begin{align*}
  \text{Denominator: } & \alpha_t \dot{\beta}_t - \dot{\alpha}_t \beta_t \\
  &= (1 - (1-\sigma_{\min})t) \cdot 1 - (-(1-\sigma_{\min})) \cdot t \\
  &= 1 - (1-\sigma_{\min})t + (1-\sigma_{\min})t \\
  &= 1.
\end{align*}

\begin{align*}
  \text{Numerator: } & \alpha_t v_{\theta}(h_t) - \dot{\alpha}_t h_t \\
  &= (1 - (1-\sigma_{\min})t) v_{\theta}(h_t) - (-(1-\sigma_{\min})) h_t \\
  &= (1 - (1-\sigma_{\min})t) v_{\theta}(h_t) + (1-\sigma_{\min}) h_t.
\end{align*}

Since $\sigma_{\min}$ is set to a negligibly small value (typically $10^{-5}$), the terms involving $\sigma_{\min}$ can be neglected, yielding the following approximation with error $O(\sigma_{\min})$:
\begin{equation}
  \hat{h}_{1|t} \approx (1 - t) v_{\theta}(h_t) + h_t.
\end{equation}

Rearranging the terms, we obtain the estimation formula presented in Eq. \ref{eq:method:h1pred}:
\begin{equation}
  \hat{h}_{1|t} = h_t + (1-t) \cdot v_{\theta}(h_t).
\end{equation}

This derivation demonstrates that under the linear Flow Matching path, the current velocity prediction provides a direct linear estimate of the final target, enabling efficient lookahead for our energy guidance mechanism. Notably, the key simplification $\alpha_t \dot{\beta}_t - \dot{\alpha}_t \beta_t = 1$ makes our estimation particularly elegant compared to general affine paths, where the denominator may depend on both $t$ and path parameters.

\section{Additional Experimental Results}
\label{sec:appendix:additional_experiments}

In this section, we provide supplementary experimental results that were omitted from the main paper due to space constraints, including the full cross-dataset generalization tables, detailed hyperparameter sensitivity analysis, and additional visualization examples.

\subsection{Full Cross-Dataset Generalization Results}
\label{sec:appendix:cross_dataset}

Table \ref{tab:cross-dataset} in the main paper presented a subset of our cross-dataset evaluation. Here, we report the complete pairwise generalization results across all five datasets. The model is trained on one source dataset and evaluated on the other four target datasets without any fine-tuning.

\begin{table*}[t]
  \centering
  \caption{EFF-Grasp achieves superior cross-dataset generalization performance.}
  \label{tab:cross-dataset}
  \resizebox{\linewidth}{!}{
  \begin{tabular}{ccc|ccc|ccc|ccc}
    \toprule
    \makebox[1.5cm][c]{Suc.6$\uparrow$} & \makebox[1.5cm][c]{Suc.1$\uparrow$} & \makebox[1.5cm][c]{Pen.$\downarrow$} & \makebox[1.5cm][c]{Suc.6$\uparrow$} & \makebox[1.5cm][c]{Suc.1$\uparrow$} & \makebox[1.5cm][c]{Pen.$\downarrow$} & \makebox[1.5cm][c]{Suc.6$\uparrow$} & \makebox[1.5cm][c]{Suc.1$\uparrow$} & \makebox[1.5cm][c]{Pen.$\downarrow$} & \makebox[1.5cm][c]{Suc.6$\uparrow$} & \makebox[1.5cm][c]{Suc.1$\uparrow$} & \makebox[1.5cm][c]{Pen.$\downarrow$} \\
    \midrule
    \multicolumn{3}{c|}{\footnotesize \textbf{DexGraspNet $\to$ UniDexGrasp}} & \multicolumn{3}{c|}{\footnotesize \textbf{DexGraspNet $\to$ DexGRAB}} & \multicolumn{3}{c|}{\footnotesize \textbf{DexGraspNet $\to$ RealDex}} & \multicolumn{3}{c}{\footnotesize \textbf{DexGraspNet $\to$ MultiDex}} \\
    70.0 & 93.5 & 19.8 & 60.8 & 91.0 & 23.8 & 41.0 & 80.1 & 18.6 & 61.0 & 93.4 & 8.0 \\
    \midrule
    \multicolumn{3}{c|}{\footnotesize \textbf{UniDexGrasp $\to$ DexGraspNet}} & \multicolumn{3}{c|}{\footnotesize \textbf{UniDexGrasp $\to$ DexGRAB}} & \multicolumn{3}{c|}{\footnotesize \textbf{UniDexGrasp $\to$ RealDex}} & \multicolumn{3}{c}{\footnotesize \textbf{UniDexGrasp $\to$ MultiDex}} \\
    67.2 & 91.5 & 21.6 & 60.6 & 91.7 & 24.1 & 42.8 & 78.8 & 20.2 & 59.4 & 94.1 & 9.5 \\
    \midrule
    \multicolumn{3}{c|}{\footnotesize \textbf{DexGRAB $\to$ DexGraspNet}} & \multicolumn{3}{c|}{\footnotesize \textbf{DexGRAB $\to$ UniDexGrasp}} & \multicolumn{3}{c|}{\footnotesize \textbf{DexGRAB $\to$ RealDex}} & \multicolumn{3}{c}{\footnotesize \textbf{DexGRAB $\to$ MultiDex}} \\
    63.4 & 89.2 & 20.4 & 63.6 & 90.3 & 18.9 & 29.0 & 58.5 & 26.1 & 55.0 & 94.7 & 9.6 \\
    \midrule
    \multicolumn{3}{c|}{\footnotesize \textbf{RealDex $\to$ DexGraspNet}} & \multicolumn{3}{c|}{\footnotesize \textbf{RealDex $\to$ UniDexGrasp}} & \multicolumn{3}{c|}{\footnotesize \textbf{RealDex $\to$ DexGRAB}} & \multicolumn{3}{c}{\footnotesize \textbf{RealDex $\to$ MultiDex}} \\
    66.4 & 90.1 & 22.0 & 69.2 & 92.6 & 19.9 & 61.0 & 91.3 & 23.5 & 51.9 & 90.0 & 8.3 \\
    \midrule
    \multicolumn{3}{c|}{\footnotesize \textbf{MultiDex $\to$ DexGraspNet}} & \multicolumn{3}{c|}{\footnotesize \textbf{MultiDex $\to$ UniDexGrasp}} & \multicolumn{3}{c|}{\footnotesize \textbf{MultiDex $\to$ DexGRAB}} & \multicolumn{3}{c}{\footnotesize \textbf{MultiDex $\to$ RealDex}} \\
    67.7 & 88.0 & 21.5 & 72.4 & 91.8 & 20.6 & 67.1 & 97.1 & 22.6 & 61.1 & 90.5 & 29.9 \\
    \bottomrule
  \end{tabular}
  }
\end{table*}

\subsection{Hyperparameter Sensitivity Analysis}
\label{sec:appendix:sensitivity}

We investigate the sensitivity of EFF-Grasp to several key hyperparameters in our energy guidance mechanism: the guidance scale $s$, the temperature $\tau$, and the weights of the energy fields. Sensitivity experiments are conducted on the MultiDex dataset. Based on the analysis below, \textbf{we identify the optimal hyperparameter setting and apply it consistently across all datasets} (DexGraspNet, UniDexGrasp, DexGRAB, RealDex, and MultiDex) in our main experiments.

\subsubsection{Effect of Guidance Scale $s$}
The guidance scale $s$ controls the strength of the physical energy field applied to the Flow Matching trajectory.
\begin{itemize}
    \item \textbf{Setup}: We vary $s \in \{0, 10, 30, 50, 100\}$ while fixing $\tau=0.05$.
    \item \textbf{Results}: As shown in Table \ref{tab:appendix-sensitivity-s}, increasing $s$ initially improves Success Rate significantly as the physical constraints are enforced. Conversely, excessively large $s$ (e.g., $s=100$) leads to trajectory instability. We choose $s=30$ as the optimal trade-off, achieving the lowest penetration depth while maintaining a competitive success rate.
\end{itemize}

\begin{table}[h]
  \centering
  \caption{Sensitivity to Guidance Scale $s$.}
  \label{tab:appendix-sensitivity-s}
  \small
  \begin{tabular}{lccccc}
    \toprule
    Guidance Scale $s$ & 0 & 10 & \textbf{30} & 50 & 100 \\
    \midrule
    Suc.6 (\%) & 65.9 & \textbf{78.8} & 77.5 & 69.1 & 26.9 \\
    Suc.1 (\%) & 95.0 & \textbf{98.8} & 97.5 & 95.6 & 50.3 \\
    Pen. (mm)    & 15.2 & 16.2 & \textbf{10.5} & 18.7 & 19.9 \\
    \bottomrule
  \end{tabular}
\end{table}

\subsubsection{Effect of Temperature $\tau$}
The temperature $\tau$ modulates the sharpness of the Boltzmann distribution in the Local Monte Carlo estimation.
\begin{itemize}
    \item \textbf{Setup}: We vary $\tau \in \{0.01, 0.05, 0.1, 1.0\}$ while fixing $s=30$.
    \item \textbf{Results}: As shown in Table \ref{tab:appendix-sensitivity-tau}, a lower temperature makes the energy landscape extremely sharp, leading to unstable gradients and sample collapse in the Monte Carlo estimation. This causes the generation process to fail frequently. Conversely, a high temperature dilutes the guidance signal. $\tau=0.05$ yields the best performance, especially in minimizing penetration depth.
\end{itemize}

\begin{table}[h]
  \centering
  \caption{Sensitivity to Temperature $\tau$.}
  \label{tab:appendix-sensitivity-tau}
  \small
  \begin{tabular}{lcccc}
    \toprule
    Temperature $\tau$ & 0.01 & \textbf{0.05} & 0.1 & 1.0 \\
    \midrule
    Suc.6 (\%) & 7.2 & 77.5 & \textbf{77.8} & 70.9 \\
    Suc.1 (\%) & 19.7 & \textbf{97.5} & \textbf{97.5} & 96.6 \\
    Pen. (mm)    & 21.0 & \textbf{10.5} & 18.9 & 14.8 \\
    \bottomrule
  \end{tabular}
  \vspace{-10pt}
\end{table}

\subsubsection{Effect of Energy Field Weights}
Finally, we analyze the impact of the weights for the three energy components: $w_{\mathrm{ERF}}$, $w_{\mathrm{SPF}}$, and $w_{\mathrm{SRF}}$. We adopt a control variate approach, varying one weight at a time while keeping the others fixed at their default value of 0.4.

\begin{table}[h]
  \centering
  \caption{Sensitivity to Energy Field Weights. We vary each weight component while fixing the others to the default value (0.4).}
  \label{tab:appendix-sensitivity-weights}
  \small
  \begin{tabular}{l c ccc}
    \toprule
    \textbf{Component} & \textbf{Weight} & Suc.6$\uparrow$ & Suc.1$\uparrow$ & Pen.$\downarrow$ \\
    \midrule
    \multirow{5}{*}{\textbf{$w_{\mathrm{ERF}}$}} & 0.1 & 76.9 & 97.5 & 19.9 \\
                                        & 0.25 & \textbf{77.5} & \textbf{98.8} & 18.7 \\
                                        & \textbf{0.4} & \textbf{77.5} & 97.5 & \textbf{10.5} \\
                                        & 0.5 & \textbf{77.5} & 97.8 & 13.1 \\
                                        & 1.0 & 70.3 & 95.3 & 16.3 \\
    \midrule
    \multirow{5}{*}{\textbf{$w_{\mathrm{SPF}}$}} & 0.1 & 72.8 & 96.8 & 12.5 \\
                                        & 0.25 & 74.1 & 97.4 & 10.9 \\
                                        & \textbf{0.4} & 77.5 & 97.5 & \textbf{10.5} \\
                                        & 0.5 & \textbf{79.1} & \textbf{98.4} & 17.6 \\
                                        & 1.0 & 62.8 & 90.3 & 19.8 \\
    \midrule
    \multirow{5}{*}{\textbf{$w_{\mathrm{SRF}}$}} & 0.1 & 77.2 & 96.9 & 12.9 \\
                                        & 0.25 & 77.1 & 96.4 & 13.1 \\
                                        & \textbf{0.4} & \textbf{77.5} & 97.5 & \textbf{10.5} \\
                                        & 0.5 & 76.3 & \textbf{98.8} & 12.5 \\
                                        & 1.0 & 76.6 & 98.1 & 12.0 \\
    \bottomrule
  \end{tabular}
\end{table}

\subsection{Additional Visualization Results}
\label{sec:appendix:visualization}

We provide more qualitative results to demonstrate the versatility of EFF-Grasp. Figure \ref{fig:appendix} showcases generated grasps across diverse object categories, highlighting the model's ability to handle complex geometries and ensure physical feasibility.

\begin{figure*}[h]
  \centering
  \includegraphics[width=\textwidth, height=1.0\textheight, keepaspectratio]{graph/appendix.png}
  \caption{Additional visualization of grasps generated by EFF-Grasp on various objects from different datasets.}
  \label{fig:appendix}
\end{figure*}

\end{document}